# Optimized Bucket Wheel Design for Asteroid Excavation

Ravi Teja Nallapu[a],
Andrew Thoesen[a], Laurence Garvie[b], Erik Asphaug[b], Jekanthan Thangavelautham*[b]

[a] *School of Energy, Matter and Transport Engineering, Arizona State University, Tempe, Arizona 85281, United States of America*
[b] *School of Earth and Space Exploration, Arizona State University, Tempe, Arizona 85281, United States of America*,
jekan@asu.edu
* Corresponding Author

**Abstract**
Current spacecraft need to launch with all of their required fuel for travel. This limits the system performance, payload capacity, and mission flexibility. One compelling alternative is to perform In-Situ Resource Utilization (ISRU) by extracting fuel from small bodies in local space such as asteroids or small satellites. Compared to the Moon or Mars, the microgravity on an asteroid demands a fraction of the energy for digging and accessing hydrated regolith just below the surface. Previous asteroid excavation efforts have focused on discrete capture events (an extension of sampling technology) or whole-asteroid capture and processing. This paper proposes an optimized bucket wheel design for surface excavation of an asteroid or small-body. Asteroid regolith is excavated and water extracted for use as rocket propellant. Our initial study focuses on system design, bucket wheel mechanisms, and capture dynamics applied to ponded materials known to exist on asteroids like Itokawa and Eros and small satellites like Phobos and Deimos. For initial evaluation of material-spacecraft dynamics and mechanics, we assume lunar-like regolith for bulk density, particle size and cohesion. We shall present our estimates for the energy balance of excavation and processing versus fuel gained. Conventional electrolysis of water is used to produce hydrogen and oxygen. It is compared with steam for propulsion and both show significant delta-v. We show that a return trip from Deimos to Earth is possible for a 12 kg craft using ISRU processed fuel.

## 1. Introduction

Asteroids and small bodies are thought to hold large reserves of resources such as water, iron, nickel, platinum and silica (Fig. 1) [1]. However, these small bodies have low-gravity that makes landing, surface mobility and surface manipulation treacherous as seen from the Hayabusa I, Soviet Union's Phobos II mission and Philae lander. A small amount of kinetic energy is sufficient for a spacecraft to take off from the surface or even achieve escape velocities. This has resulted in spacecraft such as Osiris-Rex adopting a 'touch and go' approach to obtaining samples. While such a strategy may be feasible for a sample return mission, a better solution is required for surface mining.

In this paper, we propose use of large counter-rotating bucket wheels that would roll on the surface of a small body and collect regolith. Counter rotating bucket wheels would cancel angular momentum generated by a single bucket wheel and hence this excavator can be an attachment to a hovering spacecraft. The spacecraft and bucket wheels would be powered using solar energy. The collected regolith would be transferred to a holding tank, heated and water extracted for propellant generation. Bucket wheels have been theorized to be well suited for low gravity environments such as the Moon and Mars [5, 9, 18, 22-23]. However resource rich asteroids can be in the milligravity regime and hence this poses important physical challenges in surface mobility and mining. This requires that wheels rotate slow, avoid loss of traction and that they have large mass to avoid or even minimize free-flight from the surface. A large bucket wheel would incorporate ballasts that would raise the mass of the excavator to minimize inadvertent free flight.

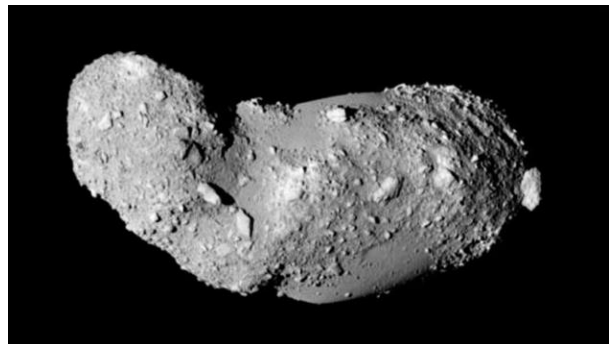

Fig. 1. Rubble pile asteroids such as Itokawa are excellent targets for future mining missions.

With sufficient mass, the bucket wheel would roll and scoop regolith to be processed. Our work shows the regolith needs to contain 5 % water content or higher for overall system feasibility.

Power is required for running the bucket wheel excavator, heating the regolith and for electrolysis. It is envisioned all three steps are done in-situ. Using this approach we develop models to design and analyse the critical design variables in the system, calculate optimal operating conditions and the main factors governing overall system performance. Our models identify important trade-offs in terms of operating time. Our work show low-power solar photovoltaic system that generates 10 kW is best suited for these applications. This 10 kW system can generate 0.4 L/hr of water, where the water content is 10 % of the regolith. Interestingly our models show higher power systems can only reduce excavation and processing time but not overall system efficiency. In addition, the impact of bucket filling efficiency is limited beyond 40 %, because the energy limiting factor is heating of regolith for water extraction and not excavation. The presented models will assist in further detailed design and refinement of asteroid surface mining concepts.

In the following sections, we present background and related work on excavators and bucket wheels (Section 2), description of the bucket wheel excavation model and analysis (Section 3), results and discussion (Section 4), followed by conclusions and future work (Section 5).

**2. Background and Related Work**

Earth based excavators like backhoes, trenchers and cranes have been very efficient on earth. They work commonly by using a brute force approach to cut through the surface. However, these devices are massive. The Bagger 293, shown in Fig. 2, is a bucket wheel excavator with a mass of 14.2 million kg.

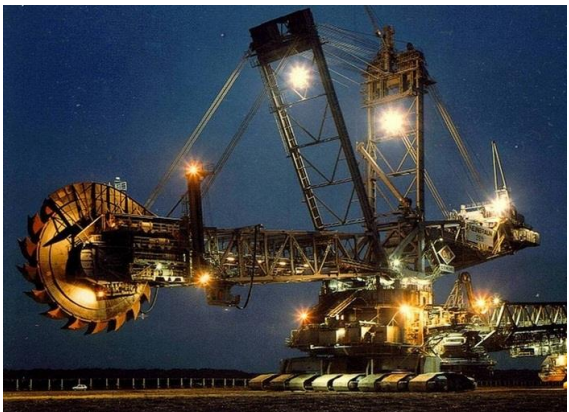

Fig. 2. Bagger 293 excavator

Therefore, despite their high efficiency, typical earth based excavators is not feasible for off-world excavation because of their launch costs. This became an important motivation to build light weight excavators that facilitate off-world excavation. NASA's Lunabotics competition have encouraged students to bring innovative ideas to build excavators suitable for mining on lunar surface. Fig. 3 and 4 show a bucket ladder and a bucket wheel excavator presented at Lunabotics 2013 [2] and 2012 [3] respectively.

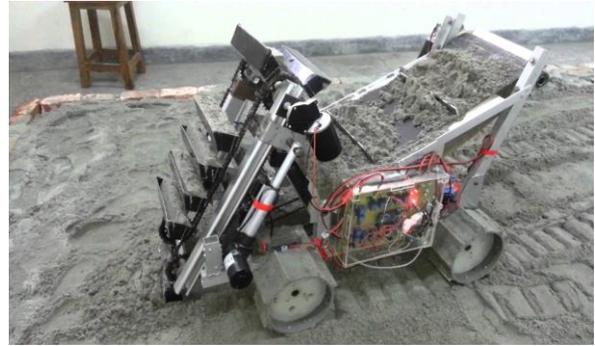

Fig. 3. Bucket ladder excavator presented at the 2013 Lunabotics competition. (Image courtesy www.cyberspaceandtime.com)

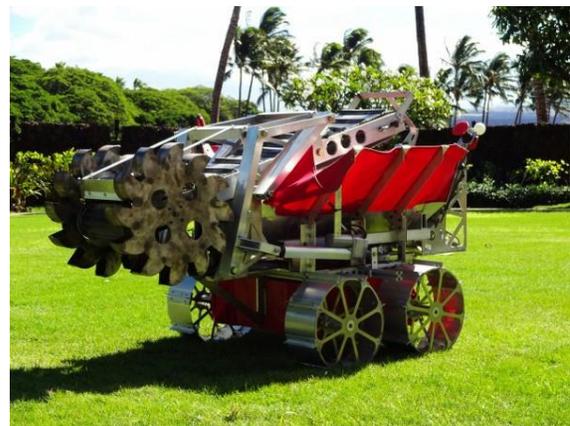

Fig. 4. Bucket wheel excavator presented at the 2012 Lunabotics competition. (Image courtesy www.adommelton.com)

Much literature exists about the performance of bucket wheels on off-world environments like Moon and Mars. Muff [4] compared different excavation mechanisms for Mars, and concluded based on his figures of merit that bucket wheels experience the least resistive forces. This was confirmed by Johnson [9]. Muller [6] did a trade study of the discrete excavation systems suitable on a lunar base, and suggested a hybrid crane line system to be an optimal solution. This study was extended by King [7] to include continuous excavators, and the work concluded continuous excavators like the bucket wheel are most desirable. A study by Johnson [8] shows that the bucket ladder has a collection rate higher than bucket wheel. However, it has been shown that the bucket ladder's chains can be easily corroded and poses a challenge in extreme environments.

Present literature focuses primarily on lunar or Mars excavation and there is limited literature on asteroid excavation. The milligravity environment of an asteroid produces severe challenges, some of which are low traction and lack of collection methods. Sonter [10] discussed the technical and economical feasibilities of mining on near earth asteroids. In his work, Sonter proposed a 3-4 tonne autonomous robot as part of a excavation mission. Similar work was done by O'Leary [11], where a feasibility study and a conceptual mission was developed to mine the surface of Phobos and Demios. A NASA Innovative and advance concepts (NIAC) report on Robotic Asteroid Prospector (RAP) suggests a solar thermal propulsion based excavator mission to supply resources and stage a vehicle platform at the Earth moon Lagrange point [12].

### 3. Bucket Wheel Design and Analysis

Staging the design of an optimal excavation system requires modelling the worksite, and optimizing the design parameters. A typical excavation and resource processing system is shown in Fig. 5.

*3.1 Modelling*

The modelling problem can be divided into the following 3 parts, including (a) worksite and mechanism modelling, (b) water extraction and (c) electrolysis/propellant generation.

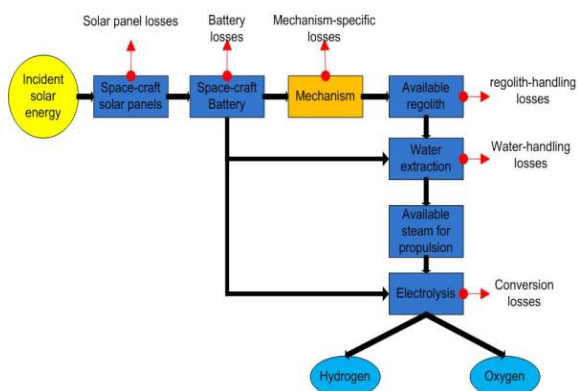

Fig. 5. Excavation and resource processing system

*3.1.1 Worksite and mechanism modelling*

Worksite, in this context describes the forces exerted by the excavation terrain. To model these forces, the Luth-Wismer model [13], which was tested by the Viking missions on Mars is used. The model was developed for pure sand (without cohesion) and for pure clay. These forces are listed below:

$$F_{sand} = \rho g w l^{1.5} \beta^{1.73} \sqrt{d} \left(\frac{d}{l \sin \beta}\right)^{0.77} \times \left[1.05 \left(\frac{d}{w}\right)^{1.11} + 1.26 \frac{v^2}{gl} + 3.91\right] \quad (1)$$

and

$$F_{clay} = \rho g w l^{1.5} \beta^{1.15} \sqrt{d} \left(\frac{d}{l \sin \beta}\right)^{1.21} \times \left[\left(\frac{11.5c}{\rho g d}\right)^{1.21} \left(\frac{2v}{3w}\right)^{0.121} \left(0.055 \left(\frac{d}{w}\right)^{0.78} + 0.065\right) + 0.64 \left(\frac{v^2}{gl}\right)\right] \quad (2)$$

Where $F_{sand}$ is the cutting force experienced in cohesion less sand, $F_{clay}$ is the cutting force experienced in a clay filled worksite, $\rho$ is the regolith density, $c$ is the sand cohesion, $g$ is the acceleration due to gravity on the surface of the worksite environment, $w$ is the width of the bucket, $l$ is the length of the cutting face of the bucket, $d$ is the penetration depth of these buckets into the regolith, $\beta$ is the angle of the buckets cutting face, and $v$ is the velocity of cutting. It is evident that the forces experienced are a dependant on the worksite, wheel, and the bucket parameters. The geometrical parameters of the bucket wheel and the buckets themselves are shown in Fig. 6 and 7.

With forces defined by Equations (1) and (2), the soil cutting forces were modelled as

$$F_{sand} + F_{clay} = F_{cut} \quad (3)$$

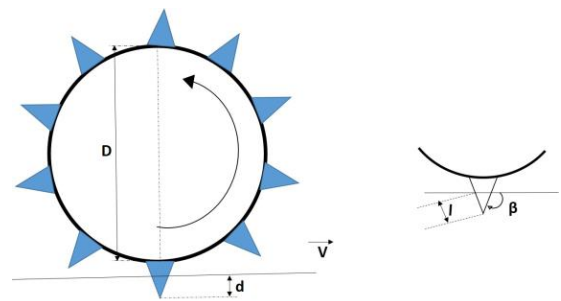

Fig. 6. Geometry of the bucket Wheel

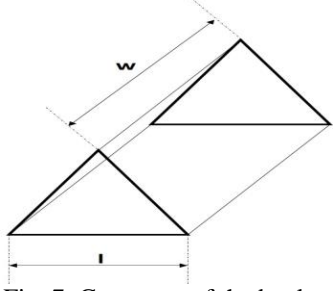

Fig. 7. Geometry of the bucket

There are high fidelity models of the excavation forces like the Balonev's [14], McKeys [15], Swick and Perumpral's models [16], however these are computationally challenging, and don't differ much from the Luth-Wismer model used [5]. To reduce the complexity of the design process, buckets presumed to be equilateral prisms similar to [18]. Hence following the same logic as [18, 19], the volume, $V$, of regolith collected in a single cut is given by:

$$V = \frac{\eta_b}{2} w l^2 \sin(60°) \quad (4)$$

where $\eta_b$ is the bucket filling efficiency. From this, the total mass collected by the excavator can be written as:

$$M_{net} = N_{rot} N_{wheel} N_{bucket} \rho V \quad (5)$$

Where $N_{rot}$ is the number of wheel rotations, $N_{wheel}$ is the number of bucket wheels of the system, and $N_{bucket}$ is the number of buckets for each wheel.

With the mass modelled, the battery power consumed for this excavation is expressed as:

$$P_{exc} = \eta_{bat} \eta_{drive} \eta_{motor} N_{rot} N_{wheel} F_{cut} v \quad (6)$$

Where $P_{exc}$ is the battery power consumed for excavation, $\eta_{bat}$ is the battery efficiency, $\eta_{drive}$ is the drivetrain efficiency which factors losses due to friction, slippage and $\eta_{motor}$ is the efficiency of the motor.

*3.1.2 Water extraction*
Starting off, the regolith is presumed to contain 5-10% water; and once the regolith is collected, the water can be extracted by heating to a 1000 °C. Factoring in water extraction efficiency of the mechanism of ($\eta_{water}$), the mass of water collected ($M_{net}$) is given as:

$$M_{water} = \eta_{water} M_{net} W_{Fr} \quad (7)$$

$W_{Fr}$ here is the normalized water content fraction in the regolith. The battery power required for this water expression is written as:

$$P_{heat} = \eta_{bat} \frac{M_{net} C_p (1000 - T_s)}{t_{heat}} \quad (8)$$

Where $P_{heat}$ is the battery power required for heating the regolith, $C_p$ is the specific heat of the regolith collected, $T s$ is the surface temperature of the regolith in Celsius, and $t_{heat}$ is the time required for heating the regolith. Fig. 8 summarises the modelling of the mechanism by showing various processes involved.

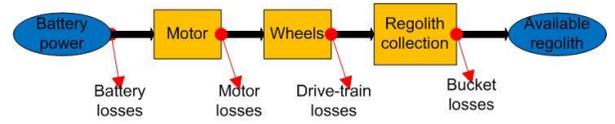

Fig. 8. Bucket wheel excavation process.

*3.1.3 Electrolysis*

One option after the water is extracted is to perform electrolysis. Water will be electrolysed to yield hydrogen and oxygen. Water contains 11.19% hydrogen, and 88.79 % oxygen by mass, therefore the mass of oxygen, and hydrogen can be calculated as follows:

$$M_{hydrogen} = 0.1119 \, \eta_H \, M_{Water}$$
$$M_{oxygen} = 0.8879 \, \eta_O \, M_{Water} \quad (9)$$

Where $\eta_H$ and $\eta_O$ are the hydrogen and oxygen extraction efficiencies, introduced to factor any extraction losses.

And the power needed for electrolysing this amount of water is given by:

$$P_{Elec} = \eta_{bat} \frac{N_{water} Q_e}{t_{Elec}} \quad (10)$$

Where $P_{Elec}$ is the battery power required for electrolysis, $N_{water}$ is the number of moles of water, $Q_e$ is the energy needed to electrolyse 1 mole of water, and is 23710 J/mole, and $t_{Elec}$ is the time required for electrolysis. Now that the 3 operations of the excavation system are modelled, the total solar power consumed can be written as:

$$P_{solar} = \eta_{solar} (P_{exc} + P_{heat} + P_{Elec}) \quad (11)$$

Where $P_{solar}$ is the solar power consumed, and $\eta_{solar}$ is the efficiency of the spacecraft-solar panels. The net operational time (t $_{Net}$) can now be defined as:

$$t_{Net} = t_{exc} + t_{heat} + t_{Elec} \quad (12)$$

Where $t_{exc}$ is the excavation time.
Finally, the following 4 performance metrics are defined to gauge the performance of the mechanism:

$$m_1 = \frac{M_{net}}{P_{solar}}$$
$$m_2 = \frac{M_{water}}{P_{solar}} \quad (13)$$
$$m_3 = \frac{M_{hydrogen}}{P_{solar}}$$
$$m_4 = \frac{M_{oxygen}}{P_{solar}}$$

### 3.2 Optimization
The optimization problem involves defining a cost function which is minimized or maximized based on a given set of constraints. Pothamsetti and Thangavelautham [20] showed that a spacecraft of net mass 12Kg needs 7.8 Kg of water to be electrolysed for net delta V of 4 km/s, enough to perform a return trip from the Mars system to Earth. Also, the maximum solar power consumed should be feasible in terms of space operations, therefore a nominal value of 10 kW was chosen as the maximum power. We later consider availability of more power and its impact on the system.

#### 3.2.1 Cost Function
A cost function (J) is defined as follows:
$$J = (P_{max} - P_{solar})^2 + 0.1(M_{req} - M_{water})^2 \quad (14)$$

Where $P_{max}$ is the maximum allowable solar power, and $M_{req}$ is the mass of water required for electrolysis. The factor 0.1 was chosen to emphasize the importance of optimizing the maximum power over the weight of water obtained.

Equation (13) will be minimized over the design space of $N_{wheel}, N_{bucket}, D, w, d, v, t_{exc}, t_{heat}, t_{Elec}$.

#### 3.2.2 Constraints
Imposing constraints on the design space is challenging because it is impossible to model all the real world constraints. A basic set of constraints were defined as shown in (15):

$$2 \leq N_{wheel} \leq 4$$
$$2 \leq N_{bucket} \leq 300$$
$$P_{Solar} \leq P_{max} \quad (15)$$
$$M_{water} \geq M_{req}$$
$$N_{bucket} \leq \frac{\pi D_{wheel}}{l}$$

#### 3.2.3 Optimizer
After the problem is formulated, solving an optimization problem requires specifying arbitrarily chosen initial conditions on the design space, this need not satisfy the constraints mentioned above. Additionally, an optimization algorithm is required to solve the problem. An optimization algorithm is one that finds a local maxima/minima. For this problem Excel's non-linear Generalized Reduced Gradient (GRG) algorithm [17] was used. One caveat in choosing the initial conditions is that it is better to start with initial conditions that are "reasonably" close to the final design one desires.

### 3.3 Design and analysis procedure
#### 3.3.1 Determining the optimal design
Running the above optimization showed that we can consistently minimize the operating time.

#### 3.3.2 Performance of the optimal design
With the optimal design, the wheel was simulated to analyse the dependence of the wheel's performance on the following parameters:
- Maximum power allowed, ($P_{max}$)
- Bucket filling efficiency, ($\eta_b$)
- Water content of regolith, ($W_{Fr}$)
- Surface temperature of regolith, ($T_s$)

## 4. Results and Discussion
This section will present and discuss the results found for the model described in Section 3. To begin with, a list of all the design requirements, surface parameters and efficiencies may be found in Table 1. As mentioned in 3.3.1, the initial number of buckets was varied, and the operation times were noted as shown in Fig 9. The optimizer alters the other operating parameters to find the shortest operating time to be 17-18 hrs to generate 7.5 L of water. This is irrespective of number of buckets. The resultant design is shown in Table 2. This further shows that there exists extensive flexibility in the design and that it is not a point design.

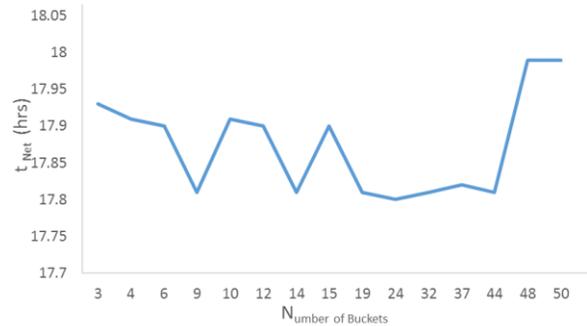

Fig. 9. Effect of number of buckets on operation times.

Table 1: Design requirement and environment.

| Design Requirements | Value |
|---|---|
| Water to be extracted (Kg) | 7.5 |
| Maximum solar power (KW) | 10 |
| **Surface Parameters** | **Value** |
| Soil density (kg/m$^3$) | 1876 |
| Acceleration due to gravity (m/s$^2$) | 0.0057 |
| Cohesion (pa) | 147 |
| Specific heat of the material (J/(Kg-Celsius)) | 1430 |
| Surface temperature of the material (Celsius) | 200 |
| Water extraction temperature (Celsius) | 1000 |
| Water content in the regolith (%) | 10 |
| **System efficiencies** | **Value** |
| Battery efficiency (%) | 75 |
| Solar panel efficiency (%) | 29 |
| Motor efficiency (%) | 70 |
| Drive train efficiency (%) | 70 |
| Bucket filling efficiency (%) | 45 |
| Water extraction efficiency (%) | 90 |
| Hydrogen extraction efficiency (%) | 90 |
| Oxygen extraction efficiency (%) | 90 |

A CAD design of a bucket wheel with 24 buckets is shown in Fig. 10. The 4 performance metrics can be seen in Fig. 11. The metric m1, which is the regolith weight collected to maximum power is about 8.33 kg/kW, which is indicative of the mass and power requirements specified.

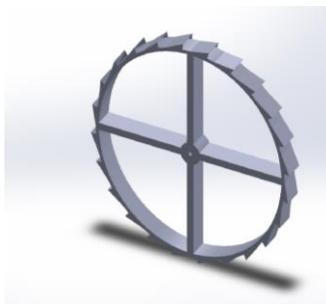

Fig. 10. CAD design of a bucket wheel.

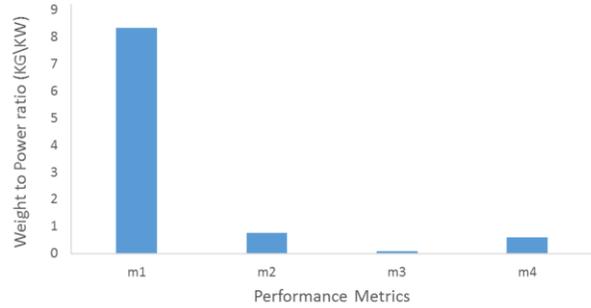

Fig. 11. Performance metrics.

The other metrics m2, m3, and m4 then scale accordingly to the amount of water in regolith, % of hydrogen in water, % of oxygen is water, respectively. Operation times were noted by varying the maximum power. As expected, increasing the maximum power decreases the operation times. This can be seen in Fig. 12. However, if one were to compare total power in to the water extracted per hr per kW (Table 3), we see maximum efficiency reached when the input power is 10 kW. When the power is increased by 5 folds, we don't see a 5 fold increase in water extracted. This suggest when there is more power available, it is better to store that power in batteries to enable continuous operation during eclipse or the additional power be used to power another bucket wheel and processing system.

Table 2: Optimal bucket wheel design parameters.

| Design Parameter | Value |
|---|---|
| # Wheels | 2 |
| # Buckets per wheel | 24 |
| Diameter of wheel(m) | 0.62 |
| Width of bucket(m) | 6.3E-02 |
| Depth of bucket(m) | 0.011 |
| Angle of buckets cut(m) | 10 |
| Cut velocity(m/s) | 0.13 |

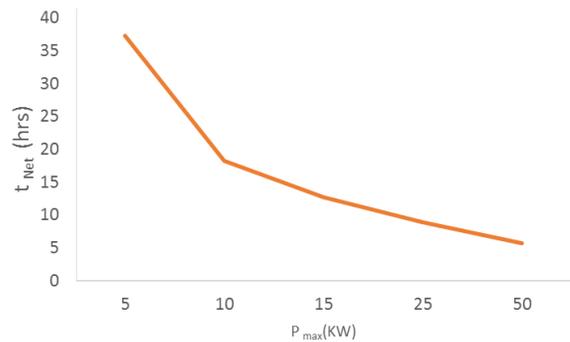

Fig. 12. Max power vs operating times.

Table 3: Overall Excavator Performance

| Power In (kW) | Water Extraction Metric (L/hr/kW) |
|---|---|
| 5 | 0.041 |
| 10 | 0.044 |
| 15 | 0.042 |
| 25 | 0.038 |
| 50 | 0.025 |

Based on these results, there is an optimal size of the power sources to bucket wheel and processing system. To scale this system up requires having many bucket wheels and processing units operating in parallel. An important environmental factor that impacts the excavator performance is the water content. Fig. 13 shows the variation of operating times based on the % of water content in the excavated regolith. Increased water content shows a linear decrease in time required as expected.

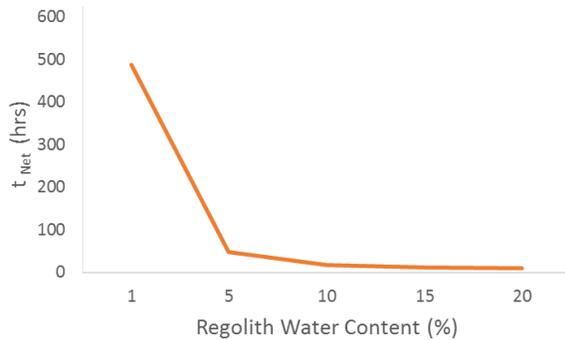

Fig. 13. Regolith water content vs operating times.

Fig. 14 shows the operating times based on the filling efficiency of the buckets. As expected, higher filling capacity of the buckets leads to lower operating times. However, beyond 40 % filling efficiency, we see minimal improvement in overall operating times. This suggests beyond a certain threshold, other factors impacting operating time as opposed to filling efficiency. One such factor is time required for heating the regolith and performing electrolysis.

We also analyse the effect of surface temperature on the overall performance of the excavation and water extraction process (Fig. 15). As temperature is lowered, we see nearly a linear increase in operating time, as heating of regolith consumes a significant portion of the solar energy.

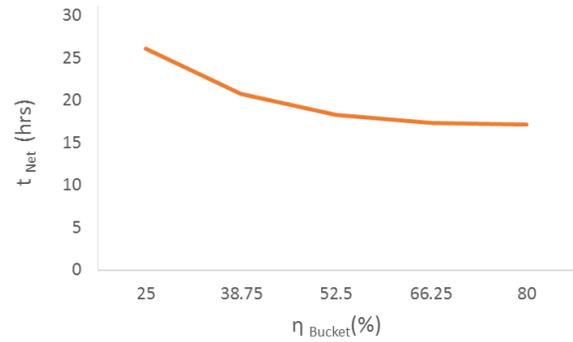

Fig. 14. Bucket filling efficiency vs operating times.

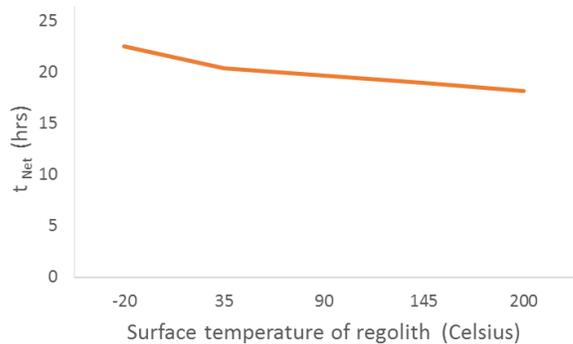

Fig. 15. Regolith surface temperature vs operating times.

Using the presented model, we have analysed the effect of various design parameters on a bucket wheel operating on an asteroid surface to collect regolith and extract water from it. Our optimization method shows robustness to various bucket wheel design parameters and finds near optimal conditions of 0.4 L/hr of water for 10 kW of power in. As would be expected, regolith water content has a significant impact on the performance of the entire system and these results presumed 10 % water content.

Our models noted that the bucket wheel filling efficiency had limited impact beyond a threshold performance. This is very good news, as surface dynamics on asteroid is one of the big unknowns particularly with a bucket wheel loosing traction on the asteroid surface or loosing contact with the asteroid surface time to time.

Using Fig. 16 from [21], 7.5 L of water generated for a 12 kg spacecraft would translate into a dry mass of 0.4. The required time for extracting the water would be 18 hours. We can achieve a delta V of 3 to 3.5 km/s conservatively speaking for a return trip to Earth from Deimos/Mars system.

Heating of the regolith was found to be the most energy intensive process and hence requires significant effort in design and operation to increase efficiency. Our work suggests solar thermal heating using carbon nanoparticles with light to heat conversion efficiencies of 80 to 99 % is well suited for this application.

From these results, we also note that there is a strong coupling between operating power and efficiency. A high power system generates more water but at the price of decreased efficiency. This suggest the excess power needs to be stored or be utilized by have multiple small, but parallel excavation operations. The use of multiple, parallel resource excavation and processing units presents both advantages and disadvantages.

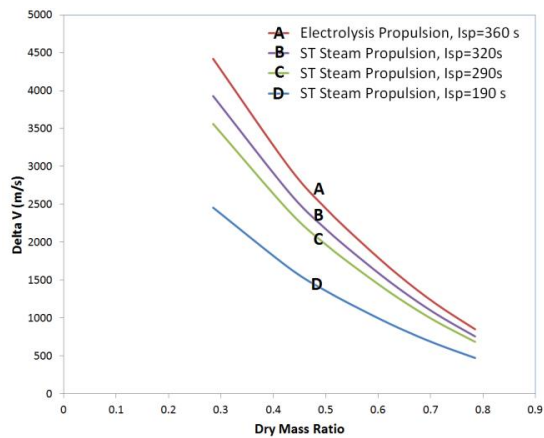

Fig. 16. Delta-v achievable for various water based propulsion systems and spacecraft dry mass ratios.

The advantage comes from being able to collect resources from multiple areas on an asteroid, which reduces risk of encountering low yield zones. In addition, having multiple parallel excavation operations makes the system immune to single point failure, down time and repairs. The challenges come from having to coordinate and control multiple resource extraction units.

Overall, the proposed design for resource extraction on an asteroid shows room for design and operation flexibility beyond a critical threshold.

## 5. Conclusions

This paper proposes an optimized bucket wheel design for asteroid or small-body excavation. Asteroid regolith is excavated and water extracted for use as rocket propellant. Our study focuses on system design, bucket wheel mechanisms, and capture dynamics applied to ponded materials known to exist on asteroids like Itokawa and Eros and small satellites like Phobos and Deimos. Our results show that a 10 kW system can generate 0.4 L/hr of water from regolith with a water content of 10 % by mass. The proposed system shows feasibility for regolith water content as low as 5 % by mass. It was found that the excavation system can tolerate low bucket filling efficiencies due to loss of traction with asteroid surface or loss of surface contact. Heating of the regolith to obtain water was found to be the most energy intensive portion of the process. Using our proposed method, we show that enough water can be generated for a 12 kg spacecraft on a return trip from Mars to Earth within 24 hours. A pathway to scale up the regolith excavation and resource extraction process is presented and this may well have implications for future human missions to the Mars system.

## References

[1] W. Bottke, A. Cellino, P. Paolicchi, R. Binzel, Asteroids III, University of Arizona Press, Tuscon, 2002.

[2] BUET Lunabotics website, 29 April,2013, http://cyberspaceandtime.com

[3] Justin, Kenneth, Bethanne, 2012 Lunabotics systems engineering paper, KSC-2012-271, Lunabotics competition, Kennedy Space Center, Florida, 2012

[4] T. C. Muff, Design, Construction, and Testing of a Small Scale Bucket Wheel Excavator for Application on the Surface of Mars, M.S. thesis, Dept. Eng., Colorado School of Mines, Golden, CO, 2003.

[5] K. Skonieczny, "Lightweight Robotic Excavation," Ph.D. dissertation, School of Computer Science, Carnegie Mellon University, Pittsburgh, PA, 2013.

[6] R. P. Mueller and R. H. King, Trade Study of Excavation Tools and Equipment for Lunar Outpost Development and ISRU, AIP Conference, Albuquerque, NM, 2008.

[7] R. H. King, M. B. Duke, and L. Johnson, Evaluation of Lunar-Regolith Excavator Concepts for a Small, ISRU Oxygen Plant, in Space Resources Roundtable VII: LEAG Conference on Lunar Exploration, League City, TX, 2005.

[8] L. L. Johnson and P. J. van Susante, "Excavation System Comparison: Bucket Wheel vs. Bucket Ladder," in Space Resources Roundtable VIII, Golden, CO, 2006.

[9] L. L. Johnson and R. H. King, Measurement of force to excavate extraterrestrial regolith with a small

bucket-wheel device, J. Terramechanics, 47(2010), 87-95.

[10] Sonter, M, J. Acta Astronautica: The Technical and Economic Feasibility of Mining the Near-Earth Asteroids. 41(1997)

[11] B. O'Leary, Asteroid Mining and the Moons of Mars, Acta Astronautica, 17 (1988).

[12] K. Zacny, P. Chu, J. Craft, M. Cohen, W. James, B. Hilscher, Asteroid mining, Proceedings of the AIAA SPACE 2013 conference and exposition, CA, San Diego, 2013

[13] T. Muff, R. King, M. Duke, Analysis of a small robot for Martian regolith excavation., AIAA Space 2001 Conference & Exposition, Albuquerque, NM, 2001, August.

[14] V. Balovnev. New methods for calculating resistance to cutting of soil, Amerind Publishing Company, 1983.

[15] E. McKyes, Soil cutting and tillage. Developments in agricultural engineering, vol. 7. Amsterdam,Elsevier; 1985.

[16] S. Swick, J. Perumpral, A model for predicting soil–tool interaction. J Terramech, 25 (1988),43–56

[17] L. Lasdon, R. Fox, M. Ratner, 1974, Nonlinear Optimization Using the Generalized Red uced Gradient Method, Rev Fr Autom Inf Rech Oper, vol. 8, pp. 73-103.

[18] K. Skonieczny, M. Delaney, D. Wettergreen, R. Whittaker, Productive Lightweight Robotic Excavation for the Moon and Mars. J. Aerosp. Eng., 27 (2014).

[19] Wettergreen, DS, Skonieczny, K & Whittaker, WL" 2010, Parameters Governing Regolith Site Work by Small Robots, pp. 1326-1333.

[20] Pothamsetti, R., Thangavelautham, J. "Photovoltaic Electrolysis Propulsion System for Interplanetary CubeSats," 2016 IEEE Aerospace Conference, 2016.

[21] S. Rabade, N. Barba, G. Liu, L. Garvie, J. Thangavelautham, "The Case for Solar Thermal Steam Propulsion System for Interplanetary Travel: Enabling Simplified ISRU Utilizing NEOs and Small Bodies," International Astronautic Congress 2016, Guadalajara, Mexico.

[22] J. Thangavelautham, K. Law, T. Fu, N. Abu El Samid, A. Smith, G. D'Eleuterio, "Autonomous Multirobot Excavation for Lunar Applications," Robotica, pp. 1-39, 2017.

[23] Thangavelautham, J., Abu El Samid, N., Grouchy, P., Earon E., Fu, T., Nagrani, N., D'Eleuterio, G.M.T., "Evolving Multirobot Excavation Controllers and Choice of Platforms Using Artificial Neural Tissue Controllers," Proceedings of the IEEE Symposium on Computational Intelligence for Robotics and Automation, 2009, DOI: 10.1109/CIRA.2009.542319